\pdfoutput=1
\documentclass[11pt]{article}

\usepackage[]{EMNLP2023}

\usepackage{times}
\usepackage{latexsym}

\usepackage[T1]{fontenc}
\usepackage[utf8]{inputenc}

\usepackage{microtype}

\usepackage{inconsolata}

\usepackage{soul}

\usepackage{multirow}

\usepackage{algorithm}
\usepackage{algpseudocode}

\usepackage{amsmath}
\usepackage{inconsolata}

\usepackage{soul}

\usepackage{multirow}

\usepackage{algorithm}
\usepackage{algpseudocode}

\usepackage{graphicx}
\usepackage{subcaption}
\usepackage{mwe}

\usepackage{amsmath}
\usepackage{mathtools}
\DeclarePairedDelimiterX{\infdivx}[2]{(}{)}{%
  #1\;\delimsize\|\;#2%
}
\newcommand{\infdiv}{D\infdivx}

\usepackage{microtype}
\usepackage{hyperref}
\usepackage{url}
\usepackage{booktabs}
\usepackage{tabularx}
\usepackage{mathtools}

\usepackage{graphicx}
\usepackage{subcaption}
\usepackage{mwe}

\title{Generalization Measures for Zero-Shot Cross-Lingual Transfer}

\author{Saksham Bassi \\
New York University\\
  \texttt{sakshambassi@nyu.edu} \\\And
  Duygu Ataman \\
New York University\\
  \texttt{ataman@nyu.edu} \\\And
   Kyunghyun Cho \\
New York University\\
  \texttt{kyunghyun.cho@nyu.edu} \\
  }

\begin{document}
\maketitle
\begin{abstract}
Building robust and reliable machine learning systems requires models with the capacity to generalize their knowledge to interpret unseen inputs with different characteristics. Traditional language model evaluation tasks lack informative metrics about model generalization, and their applicability in new settings is often measured using task and language-specific downstream performance, which is lacking in many languages and tasks. To address this gap, we explore a set of efficient and reliable measures that could aid in computing more information related to the generalization capability of language models, particularly in cross-lingual zero-shot settings. Our central hypothesis is that the sharpness of a model's loss landscape, i.e., the representation of loss values over its weight space, can indicate its generalization potential, with a flatter landscape suggesting better generalization. We propose a novel and stable algorithm to reliably compute the sharpness of a model optimum, and demonstrate its correlation with successful cross-lingual transfer. \footnote{Code: https://anonymous.4open.science/r/strikegen-7288}

\end{abstract}

\section{Introduction}

Generalization enables models to use prior knowledge to reasonably respond to previously unseen stimuli.
% A preliminary condition to be able to develop robust machine learning systems relies on measuring the capability of the system in generalizing to inputs from different domains or distributions. 
Although traditional machine learning evaluation is performed based on a preselected set of prediction or generation tasks, accuracy on many public benchmarks may often not be sufficient to extensively assess the ability to perform well in new settings. Therefore, a majority of researchers have found it worthwhile to investigate measures that could evaluate the generalization capability of models with properties, such as VC dimension \cite{Vapnik1971VCDim}, cross-entropy \cite{shannon1948mathematical}, complexity \cite{mohri2012FML} or variation in parameters during training \cite{Nagarajan2019DistInit}. 
Among these, recent findings support the smoothness in the loss curvature to correlate best with generalization capability \cite{chaudhari2019entropy,petzka2021relative,kaddour2022flat}, motivating the development of learning methods that induce smoothness in the learning trajectory such that the model becomes more robust; either through data perturbation \cite{jiang2020smart,aghajanyanbetter,wu2021r,hua2021noise,park2022consistency,zheng2021consistency,wang2021multi, huang-etal-2021-improving-zero} or by integrating the measure directly to the optimization objective \cite{izmailov2018averaging,jastrzebski2021aFIM,cha2021swad,foretsharpness,hulora,zaken2022bitfit,stickland2021regularising}. However it might often not be straightforward to compute such measures in high-dimensional feature space in a stable fashion \cite{nachum2024fantastic}.
 
As models get larger and cover more languages, the possibility of improving the applicability of NLP systems in many under-resourced languages gets more promising. An essential requirement in studying the dynamics of cross-lingual knowledge transfer is to have an evaluation methodology that can reliably measure the model's capability in generalization of knowledge under different scenarios. There is a common hypothesis that states that a model demonstrating an extended flat optimum area of low loss value surrounding the minimized loss is indicative of better generalization capability. In this work, we study the above hypothesis and present the first study to provide methods that can be used for measuring the cross-lingual generalization capability of language models. 
\begin{itemize}
    \item We pick prominent measures that were previously shown to correlate well with generalization performance \cite{jiangfantastic}, such as the Frobenius distance of the learned parameters after training \cite{Nagarajan2019DistInit}, the margin between model predictions and true labels \cite{Wei2018Margin} and sharpness in loss minima to test applicability to zero-shot cross-lingual generalization measurement \cite{keskar2017large,foretsharpness}.
    \item We also extend the formulation of state-of-the-art sharpness computation methods \cite{keskar2017large,foretsharpness} to provide a sharpness prediction algorithm such that the optimization of the parameters can converge in a more stable fashion. 
\end{itemize} 

% Our findings show that margin and sharpness in the loss optima are strong indicators of performance in cross-lingual generalization. We also find supporting evidence to the applicability of the flatness hypothesis for better generalization in the case of cross-lingual transfer, which was recently found promising in vision tasks. We observe the generalization measures and relationships can clearly capture significant differences between different learning algorithms, which would otherwise have similar down-stream task accuracy, and that these differences stay consistent across different languages. We hope our methodology provides insight and helps further explore the dynamics of cross-lingual knowledge transfer.

\section{Related work}

\textbf{Loss-landscape Minima} One of the most promising indicators of generalization capability to date seems to be related to the form of the loss landscape, in particular, the sharpness in the loss curvature. A potential reason for this fallback is traced to stochastic gradient descent (SGD) \cite{bottou2012stochastic} methods which often fall into sharp minima of the loss surface \cite{keskar2017large,chaudhari2019entropy,wang2021multi}.
Although clear conclusions on the relationship between sharpness and generalization performance, such as whether sharper \cite{dinh2017sharp} vs. flatter \cite{li2018visualizing,keskar2017large} minima would generally yield better generalization, are still due.
%\cite{izmailov2018averaging,jiang2020smart,aghajanyanbetter,wu2021r,hua2021noise,zheng2021consistency,wang2021multi,jastrzebski2021aFIM,cha2021swad,foretsharpness,stickland2021regularising,hulora,zaken2022bitfit,park2022consistency}.
The main idea behind these methods is that their objective is to explicitly find flat minima, often using stochastic averaging methods \cite{polyak1992acceleration,izmailov2018averaging}, mini-max or sharpness-aware minimization methods, which can be computed by direct formulation based on the Hessian matrix of the loss function \cite{chaudhari2019entropy,petzka2021relative} or Monte-Carlo approximations of the minimizer’s neighborhood \cite{foretsharpness,cha2021swad}.

\textbf{Adversarial optimization} Comparison of two approaches finds that for NLP tasks, mini-max methods are more competitive over averaging-based optimization \cite{kaddour2022flat}. \citet{jastrzebski2021aFIM} hypothesize that regularizing the trace of the Fisher information matrix amplifies the implicit bias of SGD, which prevents memorization. The Fisher information \cite{fisher1925theory} measures local curvature, so a smaller trace implies a flatter minimum, which gives the model more freedom to reach an optimum. Instead of explicitly minimizing the values of parameters, \citet{foretsharpness} propose minimizing both loss and sharpness while optimizing the parameters such that they lie in neighborhoods with low loss values. Perturbation is an auxiliary objective that encourages the model predictions to be similar in the vicinity of the observed training samples \cite{englesson2021generalized}, usually by penalizing the KL-divergence between the probability distribution of the perturbed and normal model. 
Perturbations can be adversarial inputs \cite{jiang2020smart} or inputs with Gaussian or uniform noise \cite{aghajanyanbetter}. To improve cross-lingual generalization, translations of the input generated by machine translation systems were used as perturbed input \cite{wang2021multi,zheng2021consistency}. Other work also has found the benefit of enforcing consistency for perturbations within the model in addition to the input distribution \cite{wu2021r,hua2021noise}.

% Multi-view subword regularization - Wang et al. First, empirically show that applying sub-word regularization during finetuning helps in cross-transfer tasks. Proposes a new regularization technique called Multi-view Subword Regularization (MVR) which essentially provides an objective function with three components: deterministic segmentation cross-entropy loss, probabilistic segmentation loss value and KL divergence between these two predictions. Deterministic Segmentation is finding the segmentation of the given word which has the highest probability Probabilistic segmentation is finding the segmentation by randomly sampling from the set of of possible segmentations MVR technique aims to reduce the predictions between deterministic and probabilistic segmentation

\section{Methodology}

In this study, we undertake the development of a methodology that could benefit an accurate assessment of the generalization capability of models for the purpose of cross-lingual knowledge transfer into under-resourced languages. This section first presents approaches to improving generalization performance and the selected measures that provide stable results for measuring zero-shot cross-lingual transfer performance.

\subsection{Sharpness-based Optimization}

We chose the following objective functions as fine-tuning methods for a given pre-trained model as a means of comparison since their main purpose is to enhance the generalization and robustness of models. Following the work of \citet{stickland2021regularising}, as the two most prominent approaches to mini-max optimization, we include Sharpness-Aware Minimization (SAM) \cite{foretsharpness} and regularization with Fisher Information Matrix (FIM) \cite{jastrzebski2021aFIM} in our evaluation study on cross-lingual generalization.We also include Multi-view Subword Regularization (MVR) as a perturbation-based optimization method \cite{wang2021multi} which induces stochasticity into the shared subword vocabulary across languages for easing cross-lingual transfer. 

\textbf{SAM} \cite{foretsharpness} works on the principle of a mini-max objective function: $\min_{w} \max_{\| \epsilon \|_{2}< \rho} L(w+\epsilon)$, which essentially means the optimizing function tries to minimize the maximum loss value in a given radius in loss landscape. Therefore, SAM states that it tries to seek "parameters lying in uniformly low-loss neighborhoods". 

\textbf{Fisher Penalty} is defined as explicitly penalizing the trace of the Fisher information matrix (FIM). \citet{jastrzebski2021aFIM}, \citet{stickland2021regularising} observed penalizing FIM during training correlates to better generalization performance. It can be written mathematically as $\frac{1}{n} \sum_{i}^{n} \nabla L(x_i, y_i)$ where $L(x_i, y_i)$ is the loss at the data point $(x_i, y_i)$.

\textbf{MVR} \cite{wang2021multi} function on the concept of consistency regularization where the divergence between the model predictions on deterministic and probabilistic segmentation inputs is minimized. The objective function is formulated as

\begin{align}
    \sum_{i=1}^{N} & \Big( -\frac{1}{2}\log p(y_i|\hat x_i) - \frac{1}{2}\log p(y_i|x_i^\prime) \\ 
    & +\lambda \infdiv{p(y_i|\hat x_i)}{p(y_i|x_i^\prime)} \Big)
\end{align}

where the first term is the model loss on \textit{deterministic segmentation} of the $i^{th}$ data sample (most probable segmentation), the second term is the model loss on \textit{probabilistic segmentation} of the $i^{th}$ data sample (random segmentation) and the third term is the KL divergence between these two output predictions. This technique influences the model to be consistent on the predictions of different input types which successively motivates the model to be more adversarially robust.

\subsection{Generalization Measures}

% Fantastic Generalization Measures and Where to Find Them - Jiang et al. Complexity measure - quantity that relates to generalization. Small complexity measure indicates smaller generalization gap. “Sharpness-based measures such as PAC-Bayesian bounds (McAllester, 1999) bounds and sharpness measure proposed by Keskar et al. (2016) perform the best overall and seem to be promising candidates for further research.” “Measures related to the optimization procedures such as the gradient noise and the speed of the optimization can be predictive of generalization.” Generalization gap `g` �� complexity measure `mu` Evaluation criteria of a complexity measure based on Kendall’s rank correlation.

%\citet{jiangfantastic} conducted an extensive study on generalization measures to find correlations between measures and model performance. It consisted of measures like norm-based metrics (based on parameter norms and distance from initial weights), flatness-based measures (sharpness metrics), and optimization-related measures. The study highlighted the success of flatness-based measures in predicting generalization in neural networks.

Our study aims to investigate which type and characteristics of methods would best correlate with better performance in generalization, in this case, zero-shot cross-lingual transfer. We are especially interested in confirming the applicability of the flatness hypothesis for cross-lingual generalization. In order to assess whether a flat optimum loss-scape region corresponds to generalization, we essentially break down the experiment to measure two things, flatness, and generalization, such that their correlation can be measured.

\citet{jiangfantastic} conducted an extensive study on image classification tasks using generalization measures such as flatness-based measures (sharpness metrics), margin and norm-based metrics (based on parameter norms and distance from initial weights) to find correlations between measures and model performance which supported the usability of measures. These measures can be useful to explore the capabilities of language models to transfer knowledge from high-resource languages to low-resource ones.

\subsubsection*{Margin}
Higher certainty in predicting the correct label leads to a model that is robust to perturbations and unseen examples. Margin is the distinction between model prediction for ground truth label and the next highest prediction probability. We use an average based margin formula defined by \citet{Wei2018Margin} to calculate margin values on the entire test set. \citet{jiangfantastic} observed that the margin was directly proportional to better generalization in the image classification tasks. Margin is

\[ \frac{1}{n} \sum_i^{n} \Big(f_{y_i}(x_i) - \max_{j \neq y_i} f_{j}(x_i)\Big) \]

where $x_i$ is the $i^{th}$ input to model, $y_i$ is the ground truth label, $f(.)$ is the model function. A larger value of the margin of a model on a given dataset would mean higher confidence in the model to predict the correct label - including unseen examples (from languages not included in fine-tuning). 
%We observe similar phenomena in our results on cross-lingual transfer.

\subsubsection*{Sharpness of optimum}
In simpler terms, we can define sharpness as the change in the model loss value at two neighboring points in the model weights plane. It can also be loosely interpreted as the inverse of the maximum radius the loss function can sustain a low loss value at the optimum. Sharpness-based measures resulted in the highest correlation with generalization in \cite{jiangfantastic}.

\citet{jiangfantastic} formulates the sharpness to be

\[ \phi = \frac{\|W - W_0\|_2^2 \log (2\omega)}{4\alpha^2} + \log \frac{m}{\sigma} + 10\]

such that $\max_{|u_i|\leq \alpha} L(f_{W+u}) < 0.1$, where $\alpha$ is the maximum radius in the model’s loss landscape possible, $W$ and $W_0$ are the models finetuned weights and model initial weights respectively, $\omega$ is the number of parameters, $m$ is the total number of observations, $\sigma$ is the standard deviation of Gaussian noise added. In this work, as we are comparing models with the same architecture (considering mBERT only), on the same dataset, we can remove the constants, and simplify the equation further for comparative analysis.
\[ \phi = \frac{\|W - W_0\|_2^2}{4\alpha^2}\]

Intuitively, if the radius of the low-loss region in the loss-landscape ($\alpha$) is small, that means the model has a higher loss value near the optimum, which would mean the landscape of the optimum is not flat. We can relate this to resulting in an unstable prediction when having perturbations in either the data or model weights. \citeauthor{jiangfantastic}'s formula didn't result in stable results for our experimental set-up which might be because the ascent steps taken to optimize the $\alpha$ value resulted in either having a large or a very small final $\alpha$. The values of $\alpha$ occurred at extreme points because the algorithm was using a binary search method and whenever optimal $\alpha$ was not found, the search algorithm stopped with the final $\alpha$ value at either of the extreme points. The correlation results of the above sharpness method are shown in Table \ref{tab:alpha-sharp-corr}.

We present an alternative definition (inclined with sharpness measure mentioned in the works of \citeauthor{keskar2017large}, and \citeauthor{foretsharpness}), $\phi_{\text{difference}}$ that removes the need to optimize $\alpha$ by calculating the difference between loss values at two points in the optimum region, formulated as

\[ \phi_{\text{difference}} = L(f_{W'}) - L(f_{W})\]

where $W'$ is $W + \epsilon$ ($\epsilon$ being Gaussian noise) and $W$ is the optimum weight parameters. The details of our definition are in Algorithm \ref{alg:diff-sharp} and performs calculation at about roughly 5-10 times faster than \citeauthor{jiangfantastic}'s algorithm for a given batch size of 8.

\begin{algorithm}
\caption{Difference-based sharpness algorithm}\label{alg:diff-sharp}
\begin{algorithmic}[1]
\State $w_0 = \text{original$\_$weight}$
\State $w = w_0 + \epsilon$ \Comment{Small noise added to avoid zero gradient}
\State $\Delta w = \nabla L(w)$
\State $w^\prime = w + n \Delta w $
\State $p=\lambda \times \|w'\|_{F}$ \Comment{$\lambda$ is small like 0.05}
\If{$\|w^\prime - w_0\| > p$}
    \State $w^\prime = w_0 + \frac{(w^\prime - w_0)}{\|w^\prime - w_0\|} \cdot p$
\EndIf
\State $\phi_{\text{difference}} = L(w^\prime) - L(w_0)$
\end{algorithmic}
\end{algorithm}

% \[
% \text{1. } w_0 = \text{original$\_$weight} \\
% \text{2. }w = w_0 + e \\

% \text{3. }\Delta w = \nabla L(w) \\

% \text{4. } w' = w + n \Delta w \\

% \text{5. } \text{proj}(w') = \begin{cases}
% w' & \text{if } ||w' - w_0|| < p \\
% w_0 + \frac{(w' - w_0)}{||w' - w_0||} \cdot p & \text{otherwise}
% \end{cases} \\

% \phi = L(w') - L(w_0)
% \]

\section{Experiments}

For comparison, we implement each Sharpness-based optimization as a fine-tuning objective on the multilingual mBERT base variant (\verb |bert-base-multilingual-cased| from huggingface) \cite{devlin2019bert} in addition to mT5 model (\verb|google/mt5-small|) \cite{xue-etal-2021-mt5}. We use a linear classification layer of size 768x3 where the output dimension is equal to the number of labels. We adopt a two-step training approach in our experiments. First, we fine-tune the model on the English language part of the XNLI dataset to optimize the model to learn the task specifically in English. Subsequently, we perform a zero-shot transfer of the fine-tuned model on the rest of the 14 languages to facilitate an evaluation of the generalization of models.

\subsection{Data, Model details, and Settings}
\label{subsec:train-det}
For this work, we used the XNLI dataset \cite{conneau2018xnli} that includes data samples from the MultiNLI dataset \cite{williams2018mnli} and their translated versions in 14 different languages (Arabic "ar", Bulgarian "bg", German "de", Greek "el", Spanish "es", French "fr", Hindi "hi", Russian "ru", Swahili "sw", Thai "th", Turkish "tr", Urdu "ur", Vietnamese "vi", Chinese "zh"). We only train the models on the English ("en") subset of the dataset. We use the data of these 14 languages only for inference and evaluation of the models.

We fine-tune pretrained mBERT models for 15 epochs each with a batch size of 32, with a learning rate of $2\times10^{-5}$, and select best checkpoint on validation. The objective function we use for the baseline model with the classification layer is the AdamW optimizer \cite{loshchilov2018adamw} with cross-entropy loss, the mBERT+FIM model has an additional loss as Fisher Penalty, the mBERT+SAM model uses the SAM optimizer and mBERT+MVR uses the MVR algorithm for fine-tuning. We use the hyperparameters and code presented in XTREME\footnote{https://github.com/google-research/xtreme} and MVR codebase\footnote{https://github.com/cindyxinyiwang/multiview-subword-regularization}. We run the models with 8 random seeds and present the average performance of these models (Figure \ref{fig:model-stability}). In Algorithm \ref{alg:diff-sharp}, the amount of Gaussian noise we add to model weights during calculating sharpness is controlled using a scale that we empirically find (among [0.001, 0.005, 0.01, 0.02]) for each model, with $n$ equal to 0.05. 

We fine-tuned the MT5 model (\verb|google/mt5-small| using Huggingface's library over 15 epochs. The XNLI dataset was processed using a function to tokenize inputs, and the optimizer utilized was Adafactor with a learning rate scheduler. Adafactor optimizer's ability to adapt learning rates is helpful with larger models like T5 in multi-lingual settings. We trained the model with a batch size of 8, accumulating gradients over 4 steps.

Additional experiments were run on PAWS-X dataset \cite{yang-etal-2019-paws} which has 7 languages: German "de", English "en", Spanish "es", French "fr", Japanese "ja", Korean "ko", Chinese "zh". We use similar experimentation of fine-tuning on english and doing a zero-shot transfer on 6 other languages as defined above for this dataset. We used Huggingface's models: mBERT (\verb|bert-base-multilingual-cased|), RoBERTa (\verb|roberta-base|), and XLM (\verb|xlm-mlm-en-2048|) using Adam optimizers.

\subsection*{Results}
To evaluate how each of the selected measures correlates with cross-lingual generalization, we first compare these measures with held-out test accuracy. In Table \ref{tab:corr-coeff}, we present the correlation coefficients (using \verb |numpy.corrcoef|) of margin vs. accuracy and sharpness vs. accuracy. We notice that having a higher margin is exceptionally correlated to achieving great performance on unseen language data. Hence, we assume the margin to indicate the generalization performance of a given model. Similarly, sharpness captures a noteworthy negative correlation with test performance.

\begin{table}[H]
\begin{center}
\begin{tabular}{lcc}
\toprule
\multicolumn{1}{c}{\bf Model}  & \multicolumn{2}{c}{\bf Correlation with Accuracy} \\
\cmidrule(lr){2-3}
 & Margin & \multicolumn{1}{c}{Sharpness} \\
\midrule
Baseline & 0.801 & -0.845 \\ 
mBERT+MVR & 0.818 & -0.793 \\ 
mBERT+SAM & 0.874 & -0.584 \\ 
mBERT+FIM & 0.954 & -0.671 \\
mT5 + Adafactor & 0.912 & -0.410 \\
\bottomrule
\end{tabular}
\end{center}
\caption{Correlation coefficients between Margin \& Test Accuracy, and Sharpness \& Test Accuracy on the XNLI dataset.}
\label{tab:corr-coeff}
\end{table}

\begin{figure}[h]
    \centering
    \includegraphics[width=0.5\textwidth]{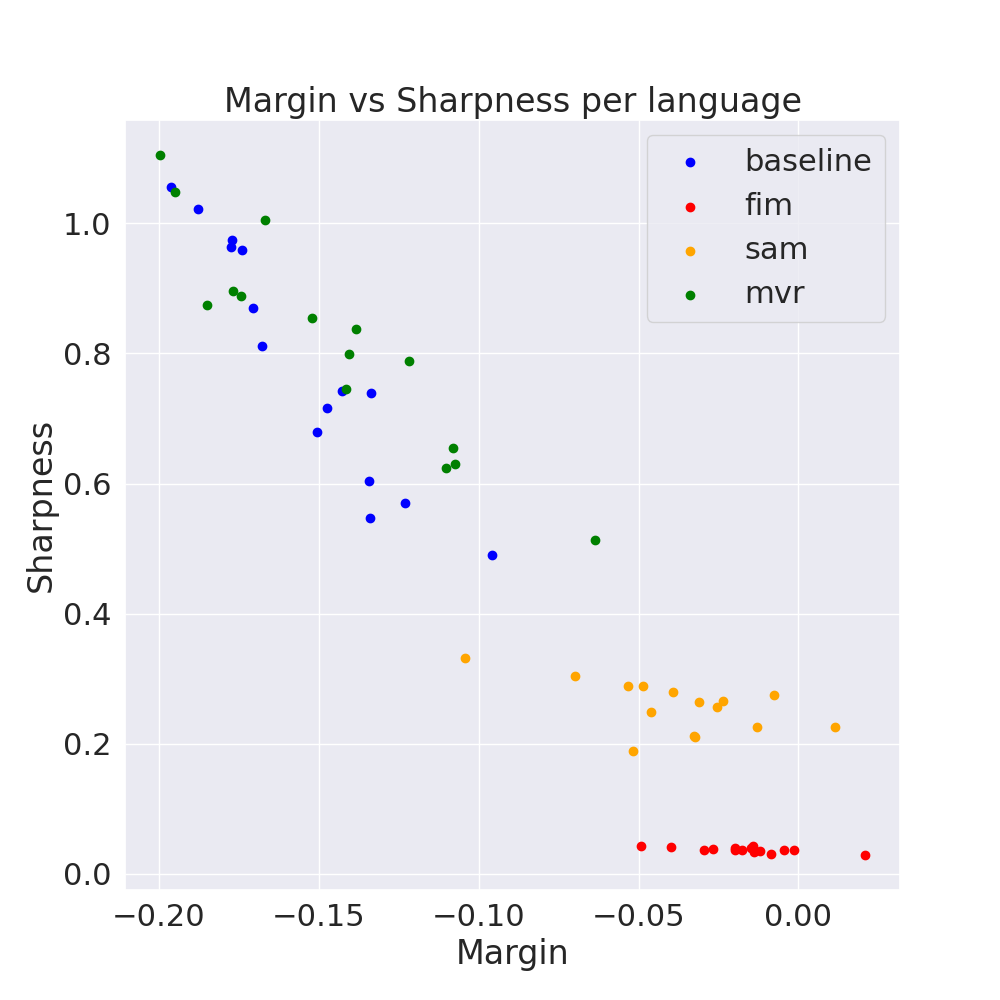}
    \caption{Scatter plot of Margin values and Sharpness ($\phi_{\text{difference}}$) values for each mBERT model (on XNLI dataset) with different objectives language-wise to show the relationship between sharpness and generalization.}
    \label{fig:margin-sharp}
\end{figure}

We notice similar results by extending our similar experimentation to Paraphrase
Identification, PAWS-X dataset \cite{yang-etal-2019-paws} with 3 different models: mBERT (\verb |bert-base-multilingual-cased|), RoBERTa (\verb |roberta-base|) \cite{roberta}, and XLM (\verb |xlm-mlm-en-2048|) \cite{xlm-CONNEAU} and analyze the validity of the flatness hypothesis, i.e. a flat optimum neighborhood would lead to a generalized model. In Figure \ref{fig:margin-sharp}, we confirm the strong relationship between Margin (indicating generalization) and Sharpness (indicating flatness) even when compared across all models and metrics, suggesting flat neighborhoods of model optimum can help in achieving higher margin values which correlate to better generalization. More findings about visualizations are in Appendix \ref{subsec:viz-results}. 

\begin{table}[H]
\begin{center}
\begin{tabular}{lcc}
\toprule
\multicolumn{1}{c}{\bf Model}  & \multicolumn{2}{c}{\bf Correlation with Accuracy} \\
\cmidrule(lr){2-3}
 & Margin & \multicolumn{1}{c}{Sharpness} \\
\midrule
mBERT & 0.998 & -0.289 \\ 
RoBERTa & 0.997 & -0.708 \\ 
XLM-R & 0.995 & -0.622 \\ 
\bottomrule
\end{tabular}
\end{center}
\caption{Correlation coefficients between Margin and Sharpness with Test Accuracy on the PAWS-X dataset.}
\label{tab:corr-coeff-paws}
\end{table}

\begin{figure}[h]
    \centering
    \includegraphics[width=0.5\textwidth]{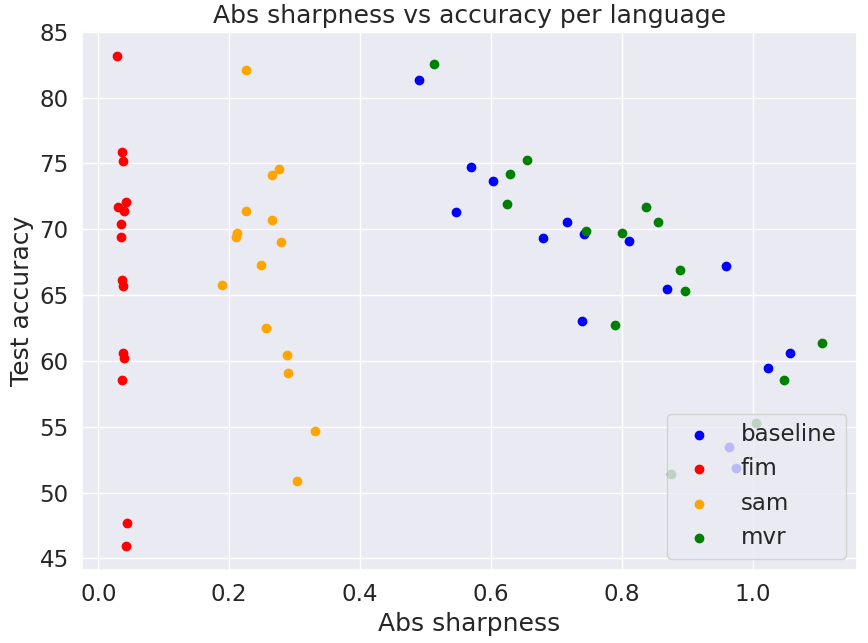}
    \caption{Scatter plot of difference-based sharpness measure with test performance for all models combined.}
    \label{fig:all-abs-sharp-acc}
\end{figure}

We can interpret sharpness as the inverse of flatness, providing us the verdict that flatness of the minimum in which the fine-tuned model is, would help the model perform better on unseen language data.
When we evaluate similar models trained with different objectives across languages, we observe that the relationships between measures are likely dependent on the optimization objective functions used during fine-tuning. In coherence with both Figure \ref{fig:margin-sharp} and \ref{fig:all-abs-sharp-acc}, overall, we see that min-max based optimization methods including FIM and SAM, have the lowest sharpness values, compared to the baseline and the regularization method MVR.

\begin{figure*}
        \centering
        \begin{subfigure}[b]{0.475\textwidth}
            \centering
            \includegraphics[width=\textwidth]{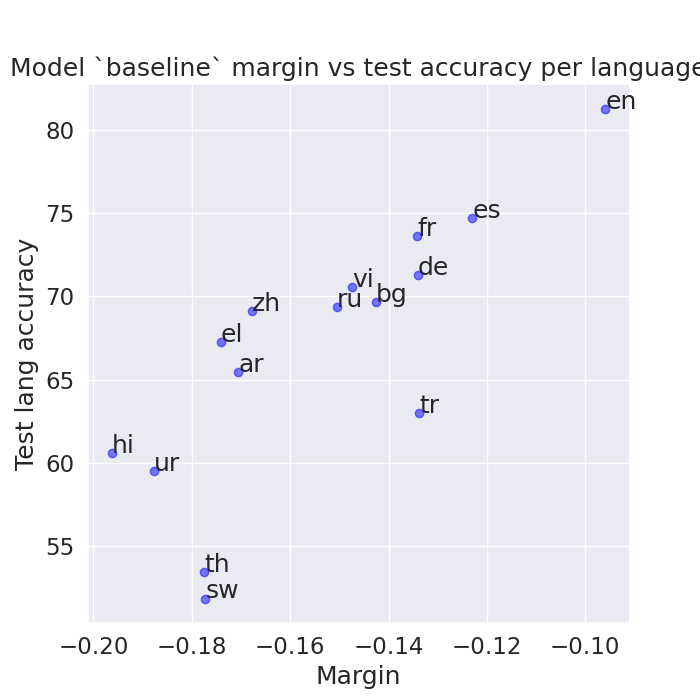}
            \caption[]%
            {{\small Correlation visualization for Baseline(mBERT + AdamW)}}    
            \label{fig:margin-sub-fig-baseline}
        \end{subfigure}
        \hfill
        \begin{subfigure}[b]{0.475\textwidth}  
            \centering 
            \includegraphics[width=\textwidth]{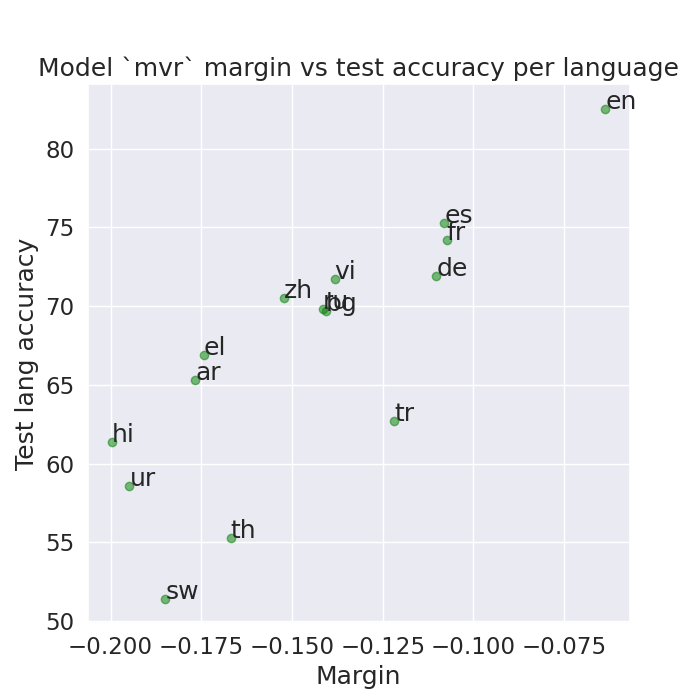}
            \caption[]%
            {{\small Correlation visualization for mBERT + MVR model}}    
            \label{fig:margin-sub-fig-mvr}
        \end{subfigure}
        \vskip\baselineskip
        \begin{subfigure}[b]{0.475\textwidth}   
            \centering 
            \includegraphics[width=\textwidth]{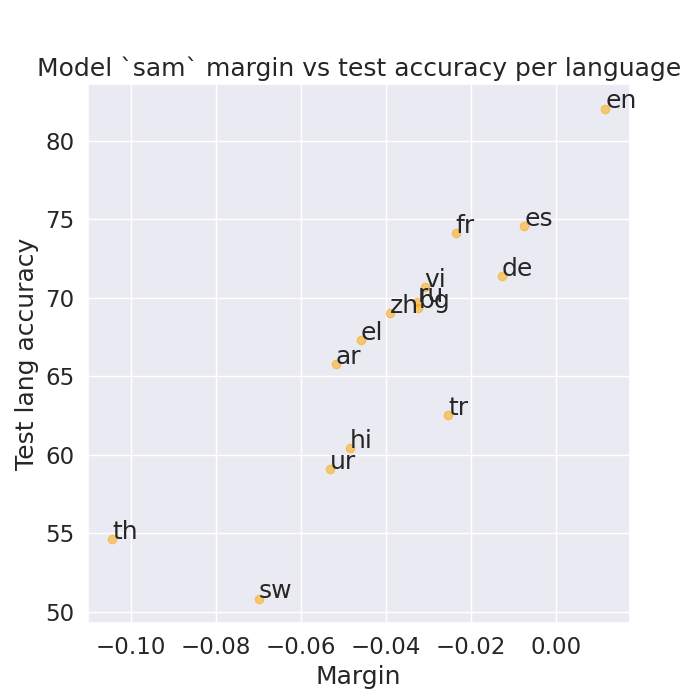}
            \caption[]%
            {{\small Correlation visualization for mBERT + SAM model}}    
            \label{fig:margin-sub-fig-sam}
        \end{subfigure}
        \hfill
        \begin{subfigure}[b]{0.475\textwidth}   
            \centering 
            \includegraphics[width=\textwidth]{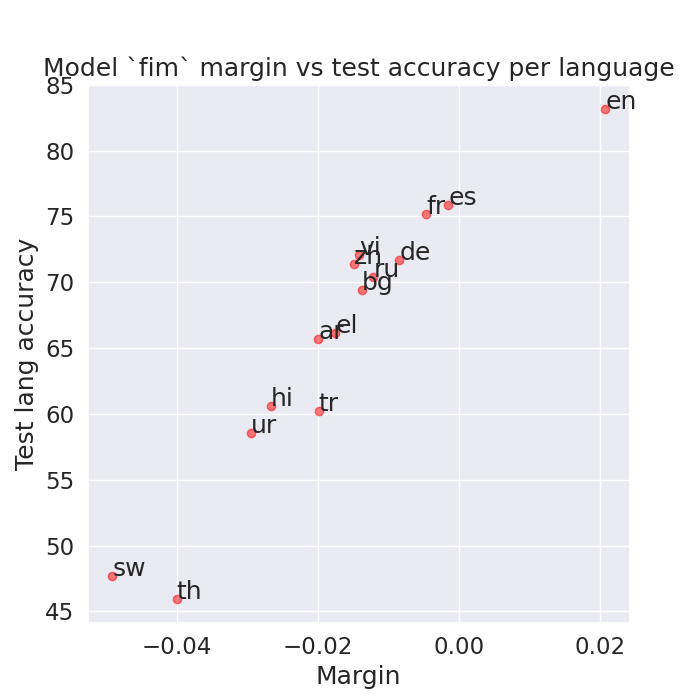}
            \caption[]%
            {{\small Correlation visualization for mBERT + FIM model}}    
            \label{fig:margin-sub-fig-fim}
        \end{subfigure}
        \caption[  ]
        {\small Scatter plots of margin of individual models and their corresponding performance on test set language-wise on XNLI dataset.} 
        \label{fig:mega-fig-margin}
    \end{figure*}

In Figure \ref{fig:mega-fig-margin}, we create scatter plots for mBERT models where in each scatter plot, we plot the model's margin based on the validation set for each language, and we plot the accuracy of that model on the test set on the XNLI dataset. We observe that the margin measure exhibits a consistent correlation with test performance across all the models analyzed.

\begin{figure*}
        \centering
        \begin{subfigure}{0.485\textwidth}
            \centering
            \includegraphics[width=\textwidth]{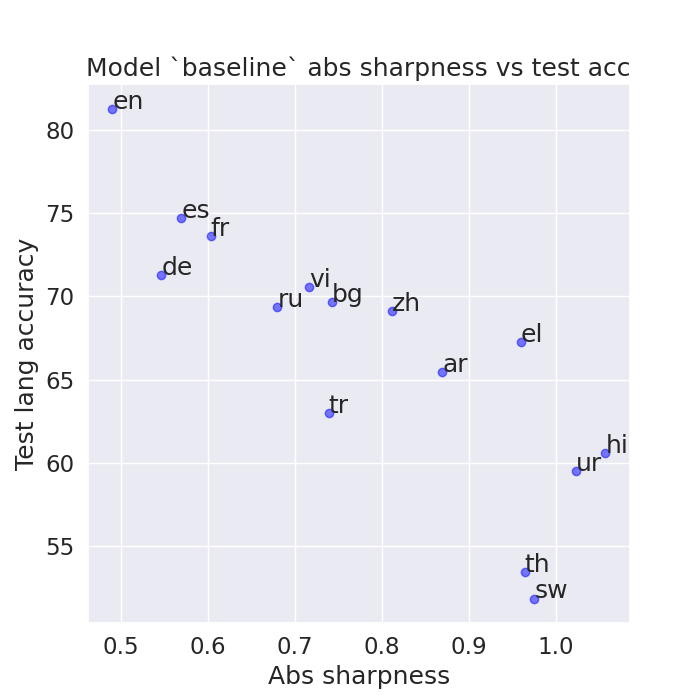}
            \caption[Network2]%
            {{\small Correlation visualization for Baseline (mBERT + AdamW)}}    
            \label{fig:sharp-sub-fig-baseline}
        \end{subfigure}
        \hfill
        \begin{subfigure}{0.485\textwidth}  
            \centering 
            \includegraphics[width=\textwidth]{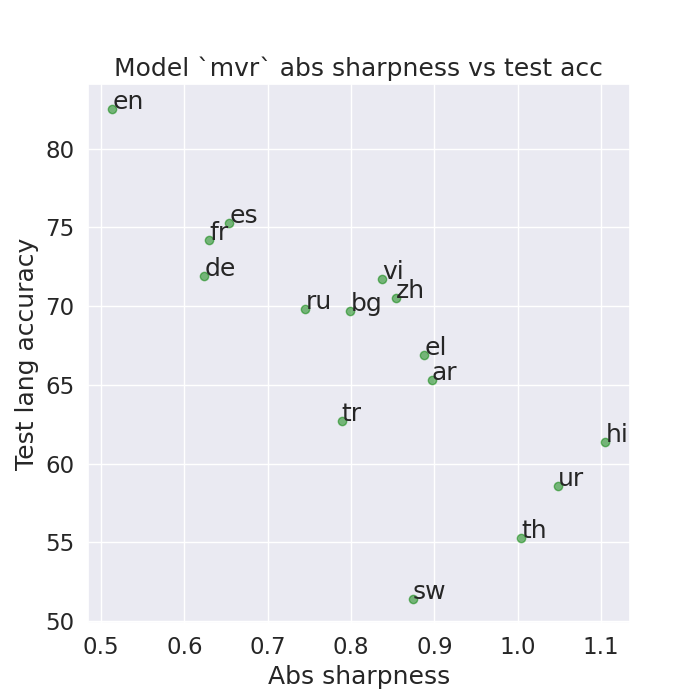}
            \caption[]%
            {{\small Correlation visualization for mBERT + MVR model}}    
            \label{fig:sharp-sub-fig-mvr}
        \end{subfigure}
        \vskip\baselineskip
        \begin{subfigure}{0.485\textwidth}   
            \centering 
            \includegraphics[width=\textwidth]{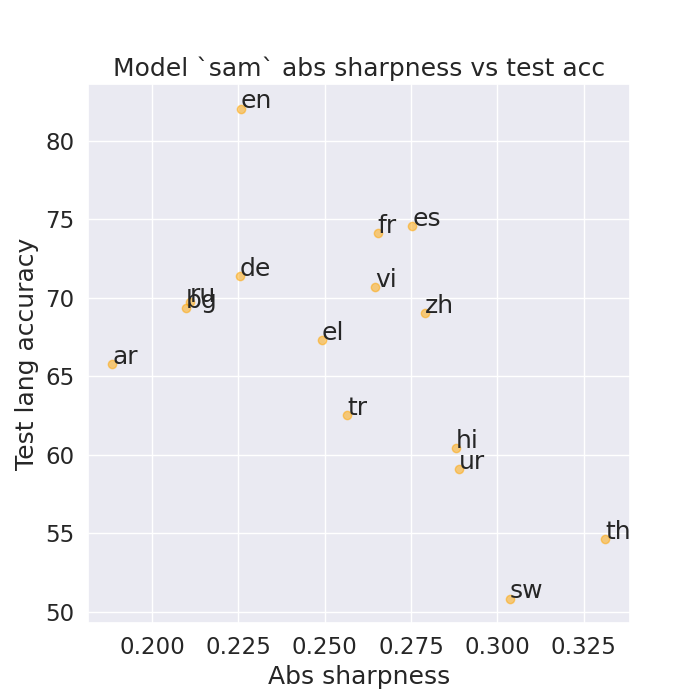}
            \caption[]%
            {{\small Correlation visualization for mBERT + SAM model}}    
            \label{fig:sharp-sub-fig-sam}
        \end{subfigure}
        \hfill
        \begin{subfigure}{0.485\textwidth}   
            \centering 
            \includegraphics[width=\textwidth]{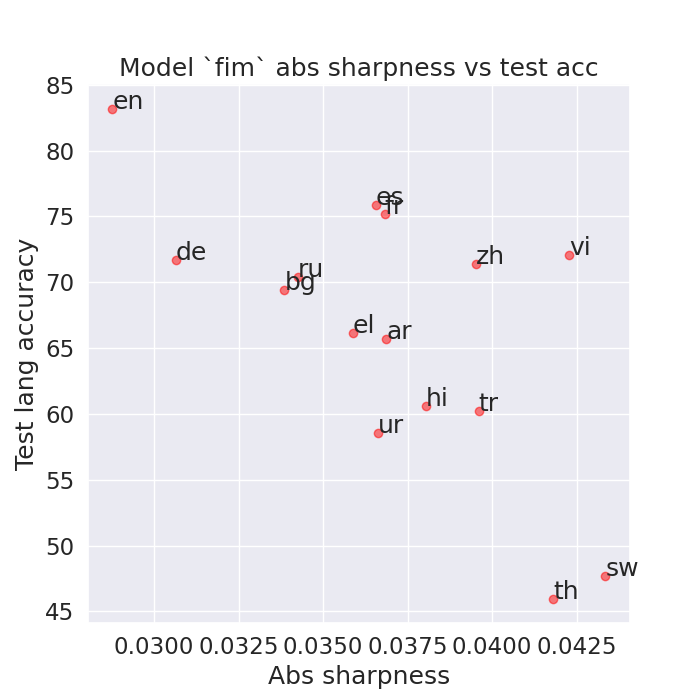}
            \caption[]%
            {{\small Correlation visualization for mBERT + FIM model}}    
            \label{fig:sharp-sub-fig-fim}
        \end{subfigure}
        \caption[ The average and standard deviation of critical parameters ]
        {\small Scatter plots of the proposed difference-based sharpness ($\phi_{\text{difference}}$) of individual models and their corresponding performance on test set language-wise on XNLI Dataset.} 
        \label{fig:mega-fig-sharp}
    \end{figure*}

As can be seen in the scatter plots for sharpness (proposed difference-based sharpness) and accuracy in Figure \ref{fig:mega-fig-sharp}, findings further indicate a negative correlation between sharpness and test performance, suggesting that lower sharpness values are associated with better generalization, represented as model performance on unseen data.

% We failed to observe a strong relationship between generalization and distance from initialization in the case of cross-lingual transfer. But, we did observe that one type of model with a high distance from initialization overall performed poorly than others.

\section{Conclusion}
Enabling cross-lingual knowledge transfer is an important step towards extending the applicability of NLP models to more languages. 
Despite recent efforts to develop better optimization methods for improving the generalization of language models in new languages or domains; 
these techniques try different types of methods to achieve higher performance such as sharpness-based minimizations, reducing gradient of loss functions, or consistency regularization. Evaluating these techniques thoroughly without a standardized methodology remains a difficult task. 
This work aims to uncover insights into how to measure cross-lingual generalization by exploring suitable measures that work well under different settings. Our experiments studying model loss landscape and parameter properties find strong relationships between the margin, sharpness in the loss minima neighborhood, and zero-shot cross-lingual downstream task performance, both on validation and test sets, supporting strong applicability to evaluate models before deploying them in new languages.

% The findings reveal a strong correlation between margin and generalization, indicating that larger margins contribute to better performance on unseen data. Additionally, sharpness exhibits a noteworthy negative correlation with generalization, highlighting the importance of flatter neighborhoods around the model optimum. 
% In addition to providing a novel set of measures for reliable evaluation of cross-lingual generalization performance, our 

\section*{Limitations}
The algorithm presented in our paper, the difference-based sharpness measure, is a great novelty for more robust sharpness computation, however, we would like  to acknowledge that a few variables in the algorithm still require tuning heuristically, including the noise scale and the multiplication coefficient required to compute the projected radius. Secondly, the mean-based margin distance is only applicable to classification tasks. Due to the limited scope of this project, we leave the development of generalization measures more suitable for generative tasks to future work.

\clearpage

\clearpage
\bibliography{anthology,custom}
\bibliographystyle{acl_natbib}

\newpage

\appendix

\section{Appendix}
\label{sec:appendix}

\subsection{Visualization results}
\label{subsec:viz-results}

\begin{figure}[h]
    \centering
    \includegraphics[width=0.5\textwidth]{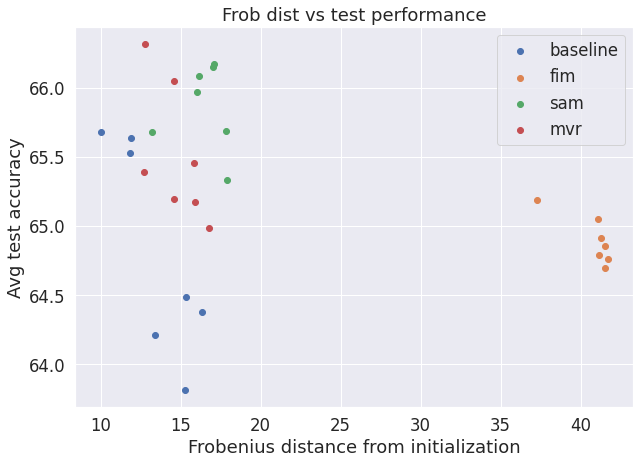}
    \caption{Scatter plot of Frobenius distance from initialization and Test accuracy for each model type (trained multiple times independenty).}
    \label{fig:frob-dist-perf}
\end{figure}

\begin{figure}
    \centering
    \includegraphics[width=0.47\textwidth]{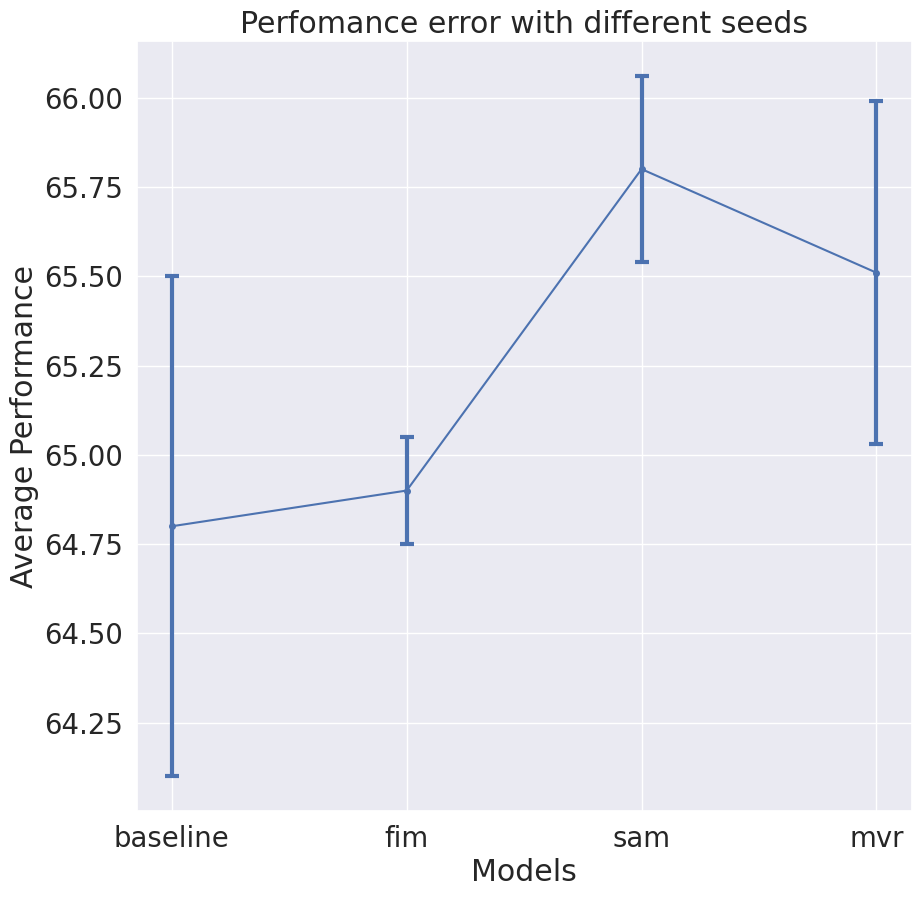}
    \caption{Average test performance (and deviations) of models when trained multiple times with different seeds.}
    \label{fig:model-stability}
\end{figure}

Previous work \cite{Nagarajan2019DistInit, jiangfantastic} suggests that a lower Frobenius distance from initialization would lead to better generalization. As Figure \ref{fig:frob-dist-perf} shows, we fail to observe a strong direct relationship between generalization and Frobenius distance from initialization. However, the model trained with Fisher Penalty as an additional objective function that has a high distance from initialization overall performed poorly than others. We also see that models trained with Fisher Penalty, SAM, and MVR optimizers tend to be more stable than the baseline model, with Fisher Penalty resulting in the most stable model when trained multiple times (with different seeds, see Figure \ref{fig:model-stability}), and SAM achieving generally the best average zero-shot task accuracy.

\subsection{Additional experiments}
\label{subsec:add-exp}

To compare all models together, using the difference-based sharpness measure, on language-wise performance, we observe it is dependent on the learning algorithms used during training in Figure \ref{fig:all-abs-sharp-acc}.

% \begin{table}
% \begin{center}
% \begin{tabular}{ll}
% \toprule
% \multicolumn{1}{c}{\bf Model}  &\multicolumn{1}{c}{\bf Correlation coefficient of $\alpha$ sharpness with accuracy} \\
% \midrule
% Baseline & 0.249 \\ 
% mBERT+MVR & -0.471 \\ 
% mBERT+SAM & -0.166 \\ 
% mBERT+FIM & -0.440 \\ 
% \bottomrule
% \end{tabular}
% \end{center}
% \caption{Correlation coefficients between $\alpha$-Sharpness \cite{jiangfantastic} \& Test Accuracy on the XNLI dataset.}
% \label{tab:alpha-sharp-corr}
% \end{table}

\begin{table}[ht]
\centering
\begin{tabularx}{\linewidth}{lX}
\toprule
\multicolumn{1}{c}{\bf Model}  &\multicolumn{1}{c}{\bf Correlation coefficient of} \\
                               &\multicolumn{1}{c}{\bf $\alpha$-sharpness with accuracy} \\
\midrule
Baseline & 0.249 \\ 
mBERT+MVR & -0.471 \\ 
mBERT+SAM & -0.166 \\ 
mBERT+FIM & -0.440 \\ 
\bottomrule
\end{tabularx}
\caption{Correlation coefficients between $\alpha$-sharpness \cite{jiangfantastic} \& Test Accuracy on the XNLI dataset.}
\label{tab:alpha-sharp-corr}
\end{table}

\begin{figure*}
        \centering
        \begin{subfigure}[b]{0.485\textwidth}
            \centering
            \includegraphics[width=\textwidth]{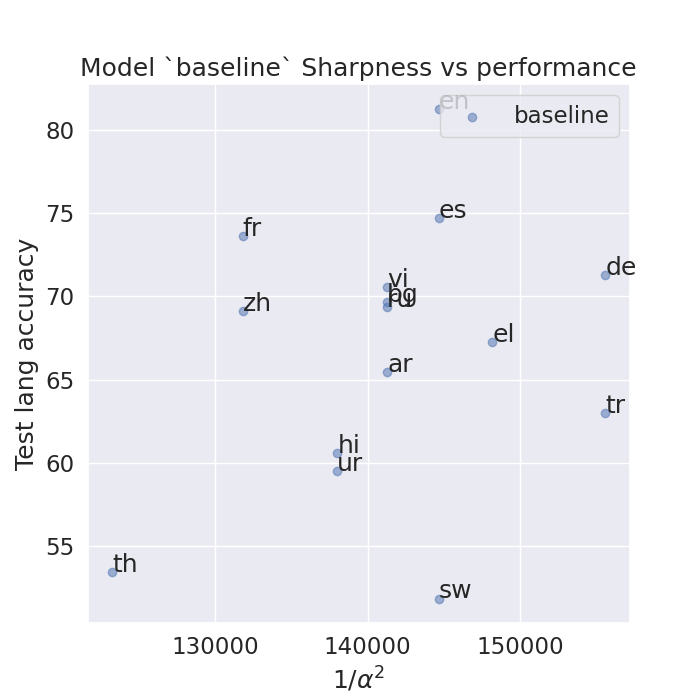}
            \caption[]%
            {{\small Correlation visualization for Baseline (mBERT + AdamW)}}    
            \label{fig:sharp-sub-fig-baseline}
        \end{subfigure}
        \hfill
        \begin{subfigure}[b]{0.485\textwidth}  
            \centering 
            \includegraphics[width=\textwidth]{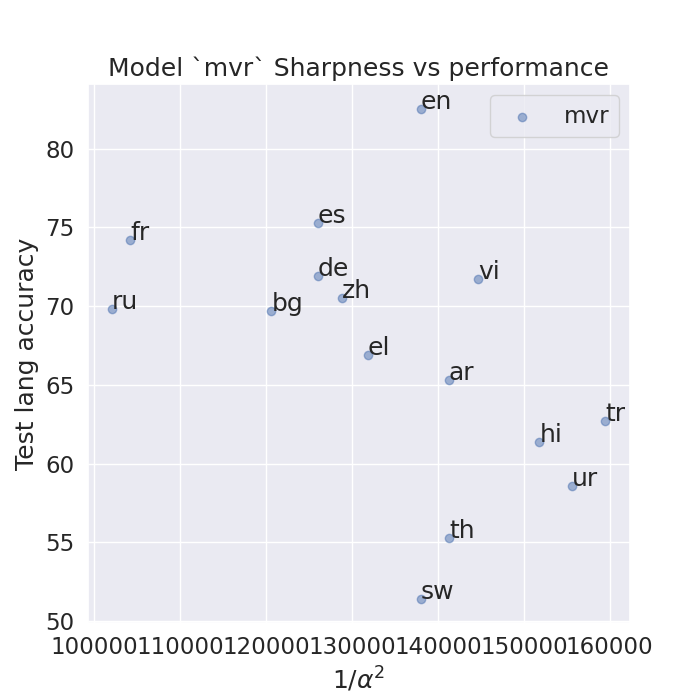}
            \caption[]%
            {{\small Correlation visualization for mBERT + MVR model}}    
            \label{fig:sharp-sub-fig-mvr}
        \end{subfigure}
        \vskip\baselineskip
        \begin{subfigure}[b]{0.485\textwidth}   
            \centering 
            \includegraphics[width=\textwidth]{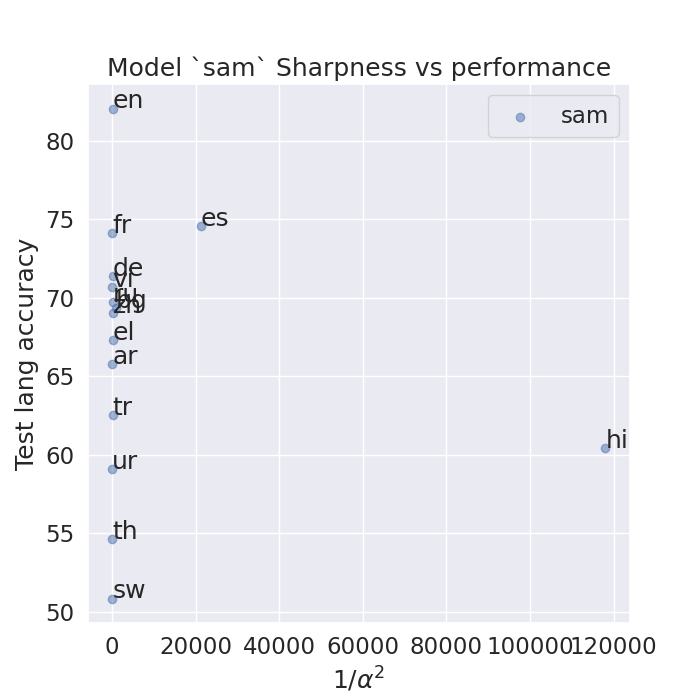}
            \caption[]%
            {{\small Correlation visualization for mBERT + SAM model}}    
            \label{fig:sharp-sub-fig-sam}
        \end{subfigure}
        \hfill
        \begin{subfigure}[b]{0.485\textwidth}   
            \centering 
            \includegraphics[width=\textwidth]{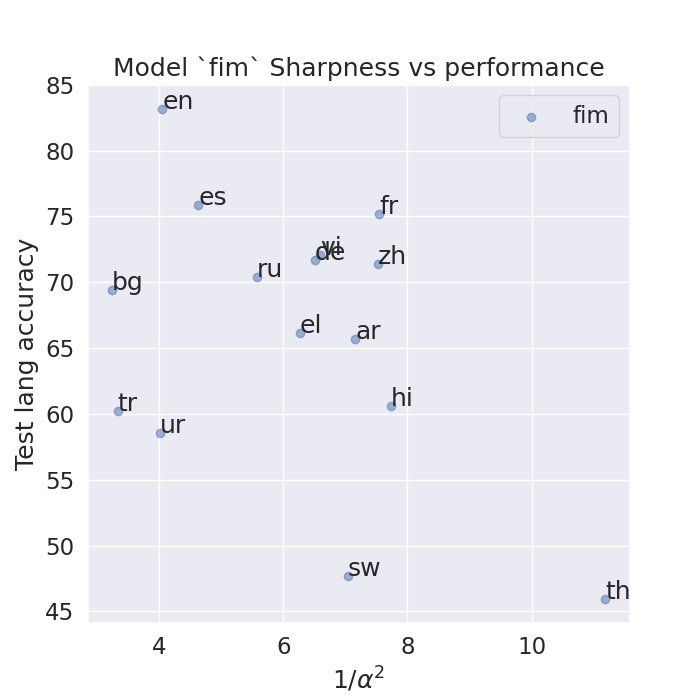}
            \caption[]%
            {{\small Correlation visualization for mBERT + FIM model}}    
            \label{fig:sharp-sub-fig-fim}
        \end{subfigure}
        \caption[ The average and standard deviation of critical parameters ]
        {\small Scatter plots of \citet{jiangfantastic} based $\alpha$-sharpness measure (we are only considering $\frac{1}{\alpha^2}$ here) of individual models and their corresponding performance on test set language-wise.} 
        \label{fig:mega-fig-alpha-sharp}
    \end{figure*}

We used the \citeauthor{jiangfantastic}'s $\alpha$-based sharpness algorithm for the experiment and optimized the threshold loss values for our experimental setting. The results of the correlation coefficient (using \verb|numpy.corrcoef|) for $\alpha$-based sharpness and test accuracy are shown in Table \ref{tab:alpha-sharp-corr} and Figure \ref{fig:mega-fig-alpha-sharp}. We notice that $\alpha$-based sharpness values occur at extreme points (for example, for mBERT+FIM model, sharpness values are low whereas for the Baseline or mBERT+MVR model, sharpness values are much larger). Apart from being a computationally expensive algorithm, we failed to see a strong relationship of $\alpha$-based sharpness with performance in Baseline and mBERT+SAM models.

\end{document}